# A Principled Approach Towards Symbolic
# Geometric Constraint Satisfaction

**Sanjay Bhansali**                                                    BHANSALI@EECS.WSU.EDU
*School of EECS, Washington State University*
*Pullman, WA 99164-2752*

**Glenn A. Kramer**                                                         GAK@EIT.COM
*Enterprise Integration Technologies, 800 El Camino Real*
*Menlo Park, CA 94025*

**Tim J. Hoar**                                                    TIMHOAR@MICROSOFT.COM
*Microsoft Corporation*
*One Microsoft Way, 2/2069*
*Redmond, WA 98052*

## Abstract

An important problem in geometric reasoning is to find the configuration of a collection of geometric bodies so as to satisfy a set of given constraints. Recently, it has been suggested that this problem can be solved efficiently by *symbolically reasoning about geometry*. This approach, called *degrees of freedom analysis*, employs a set of specialized routines called *plan fragments* that specify how to change the configuration of a set of bodies to satisfy a new constraint while preserving existing constraints. A potential drawback, which limits the scalability of this approach, is concerned with the difficulty of writing plan fragments. In this paper we address this limitation by showing how these plan fragments can be automatically synthesized using first principles about geometric bodies, actions, and topology.

## 1. Introduction

An important problem in geometric reasoning is the following: given a collection of geometric bodies, called *geoms*, and a set of constraints between them, find a *configuration* – i.e., position, orientation, and dimension – of the geoms that satisfies all the constraints. Solving this problem is an integral task for many applications like constraint-based sketching and design, geometric modeling for computer-aided design, kinematics analysis of robots and other mechanisms (Hartenberg & Denavit, 1964), and describing mechanical assemblies.

General purpose constraint satisfaction techniques are not well suited for solving constraint problems involving complicated geometry. Such techniques represent geoms and constraints as algebraic equations, whose real solutions yield the numerical values describing the desired configuration of the geoms. Such equation sets are highly non-linear and highly coupled and in the general case require iterative numerical solutions techniques. Iterative numerical techniques are not particularly efficient and can have problems with stability and robustness (Press, Flannery, Teukolsky & Vetterling, 1986). For many tasks (e.g., simulation and optimization of mechanical devices) the same equations are solved repeatedly which makes a compiled solution desirable. In theory, symbolic manipulation of equations can often yield a non-iterative, closed form solution. Once found, such a closed-form solution can be executed very efficiently.

However, the computational intractability of symbolic algebraic solution of the equations renders this approach impractical (Kramer, 1992; Liu & Popplestone, 1990).

In earlier work Kramer describes a system called GCE that uses an alternative approach called *degrees of freedom analysis* (1992, 1993). This approach is based on symbolic reasoning about *geometry*, rather than equations, and was shown to be more efficient than systems based on algebraic equation solvers. The approach uses two models. A symbolic geometric model is used to reason symbolically about how to assemble the geoms so as to satisfy the constraints incrementally. The "assembly plan" thus developed is used to guide the solution of the complex nonlinear equations - derived from the second, numerical model - in a highly decoupled, stylized manner.

The GCE system was used to analyze problems in the domain of kinematics and was shown to perform kinematics simulation of complex mechanisms (including a Stirling engine, an elevator door mechanism, and a sofa-bed mechanism) much more efficiently than pure numerical solvers (Kramer, 1992). The GCE has subsequently been integrated in a commercial system called Bravo™ by Applicon where it is used to drive the 2D sketcher (Brown-Associates, 1993). Several academic systems are currently using the degrees of freedom analysis for other applications like assembly modeling (Anantha, Kramer & Crawford, 1992), editing and animating planar linkages (Brunkhart, 1994), and feature-based design (Salomons, 1994; Shah & Rogers, 1993).

GCE employs a set of specialized routines called *plan fragments* to create the assembly plan. A plan fragment specifies how to change the configuration of a geom using a fixed set of operators and the available degrees of freedom, so that a new constraint is satisfied while preserving all prior constraints on the geom. The assembly plan is completed when all constraints have been satisfied or the degrees of freedom is reduced to zero. This approach is canonical: the constraints may be satisfied in any order; the final status of the geom in terms of remaining degrees of freedom is the same (p. 80-81, Kramer, 1992). The algorithm for finding the assembly procedure has a time complexity of $O(cg)$ where $c$ is the number of constraints and $g$ is the number of geoms (p. 139, Kramer, 1992).

Since the crux of problem-solving is taken care of by the plan fragments, the success of the approach depends on one's ability to construct a complete set of plan fragments meeting the canonical specification. The number of plan fragments needed grows geometrically as the number of geoms and constraints between them increase. Worse, the complexity of the plan fragments increases exponentially since the various constraints interact in subtle ways creating a large number of special cases that need to be individually handled. This is potentially a serious limitation in extending the degrees of freedom approach. In this paper we address this problem by showing how plan fragments can be automatically generated using first principles about geoms, actions, and topology.

Our approach is based on planning. Plan fragment generation can be reduced to a planning problem by considering the various geoms and the invariants on them as describing a *state*. Operators are actions, such as *rotate*, that can change the configuration of geoms, thereby violating or achieving some constraint. An *initial* state is specified by the set of existing invariants on a geom and a *final* state by the additional constraints to be satisfied. A plan is a sequence of actions that when applied to the initial state achieves the final state.

With this formulation, one could presumably use a classical planner, such as STRIPS (Fikes & Nilsson, 1971), to automatically generate a plan-fragment. However, the operators in this domain are parametric operators with a real-valued domain. Thus, the search space consists of an infinite number of states. Even if the real-valued domain is discretized by considering real-valued intervals there is still a very large search space and finding a plan that satisfies the specified constraints would be an intractable problem. Our approach uses *loci information* (representing a set of points that satisfy some constraints) to reason about the effects of various operators and thus reduces the search problem to a problem in topology, involving reasoning about the intersection of various loci.

An issue to be faced in using a conventional planner is the *frame problem*: how to determine what properties or relationships do not change as a result of an action. A typical solution is to use the assumption: an action does not modify any property or relationship unless explicitly stated as an effect of the action. Such an approach works well if one knows *a priori* all possible constraints or invariants that might be of interest and relatively few constraints get affected by each action - which is not true in our case. We use a novel scheme for representing effects of actions. It is based on reifying (i.e., treating as first class objects) actions in addition to geometric entities and invariant types. We associate, with each pair of geom and invariants, a set of actions that can be used to achieve or preserve that invariant for that geom. Whenever a new geom or invariant type is introduced the corresponding rules for actions that can achieve/preserve the invariants have to be added. Since there are many more invariant types than actions in this domain, this scheme results in simpler rules. Borgida, Mylopoulos & Reiter (1993) propose a similar approach for reasoning with program specifications. A unique feature of our work is the use of geometric-specific matching rules to determine when two or more general actions that achieve/preserve different constraints can be reformulated to a less general action.

Another shortcoming of using a conventional planner is the difficulty of representing conditional effects of operators. In GCE an operation's effect depends on the type of geom as well as the particular geometry. For example, the action of translating a body to the intersection of two lines on a plane would normally reduce the body's translational degrees of freedom to zero; however, if the two lines happen to coincide then the body still retains one degree of translational freedom and if the two lines are parallel but do not coincide then the action fails. Such situations are called *degeneracies*. One approach to handling degeneracies is to use a reactive planner that dynamically revises its plan at run-time. However, this could result in unacceptable performance in many real-time applications. Our approach makes it possible to pre-compile all potential degeneracies in the plan. We achieve this by dividing the planning algorithm into two phases. In the first phase a skeletal plan is generated that works in the normal case and in the second phase, this skeletal plan is refined to take care of singularities and degeneracies. The approach is similar to the idea of refining skeletal plans in MOLGEN (Friedland, 1979) and the idea of critics in HACKER (Sussman, 1975) to fix known bugs in a plan. However, the skeletal plan refinement in MOLGEN essentially consisted of instantiating a partial plan to work for specific conditions, whereas in our method a complete plan which works for a normal case is extended to handle special conditions like degeneracies and singularities.

## 1.1 A Plan Fragment Example.

We will use a simple example of a plan fragment specification to illustrate our approach. Domains such as mechanical CAD and computer-based sketching rely heavily on complex combinations of relatively simple geometric elements, such as points, lines, and circles and a small collection of constraints such as coincidence, tangency, and parallelism. Figure 1 illustrates some fairly complex mechanisms (all implemented in GCE) using simple geoms and constraints.

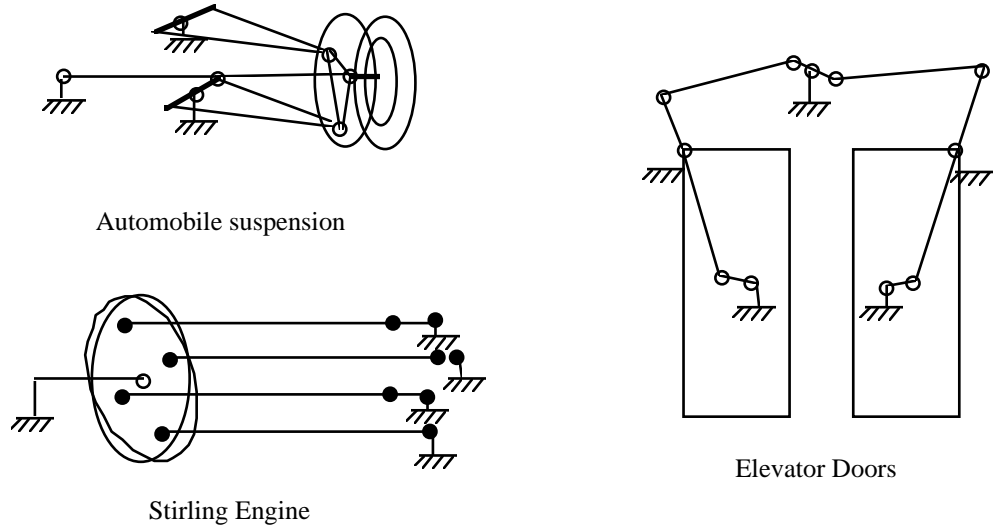

Automobile suspension

Stirling Engine

Elevator Doors

Figure 1. Modeling complex mechanisms using simple geoms and constraints. All the constraints needed to model the joints in the above mechanisms are solvable using the degrees of freedom approach.

Our example problem is illustrated in Figure 2 and is specified as follows:

**Geom-type:** circle
**Name:** $c
**Invariants:** (fixed-distance-line $c $L1 $dist1 BIAS_COUNTERCLOCKWISE)
**To-be-achieved:** (fixed-distance-line $c $L2 $dist2 BIAS_CLOCKWISE)

In this example, a variable-radius circle $c$[1] has a prior constraint specifying that the circle is at a fixed distance *$dist1* to the left of a fixed line *$L1* (or alternatively, that a line drawn parallel to *$L1* at a distance $dist1 from the center of *$c* is tangent in a counterclockwise direction to the circle). The new constraint to be satisfied is that the circle be at a fixed distance *$dist2* to the right of another fixed line *$L2*.

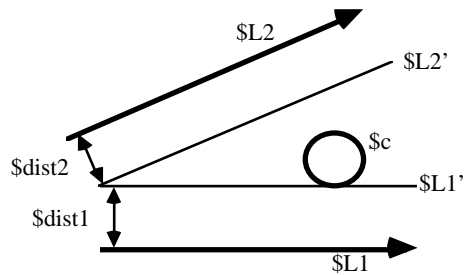

Figure 2. Example problem (initial state)

---

[1]We use the following conventions: symbols preceded by $ represent constants, symbols preceded by ? represent variables, expressions of the form (>> parent subpart) denote the subpart of a compound term, parent.

To solve this problem, three different plans can be used: (a) translate the circle from its current position to a position such that it touches the two lines $L2'$ and $L1'$ shown in the figure (b) scale the circle while keeping its point of contact with $L1'$ fixed, so that it touches $L2'$ (c) scale and translate the circle so that it touches both $L2'$ and $L1'$.

Each of the above action sequences constitute one plan fragment that can be used in the above situation and would be available to GCE from a plan-fragment library. Note that some of the plan fragments would not be applicable in certain situations. For example, if $L1$ and $L2$ are parallel, then a single translation can never achieve both the constraints, and plan-fragment (a) would not be applicable. In this paper we will show how each of the plan-fragments can be automatically synthesized by reasoning from more fundamental principles.

The rest of the paper is organized as follows: Section 2 gives an architectural overview of the system built to synthesize plan fragments automatically with a detailed description of the various components. Section 3 illustrates the plan fragment synthesis process using the example of Figure 2. Section 4 describes the results from the current implementation of the system. Section 5 relates our approach to other work in geometric constraint satisfaction. Section 6 summarizes the main results and suggests future extensions for this work.

## 2. Overview of System Architecture

Figure 3 gives an overview of the architecture of our system showing the various knowledge components and the plan generation process. The knowledge represented in the system is broadly categorized into a *Geom knowledge-base* that contains knowledge specific to particular geometric entities and a *Geometry knowledge-base* that is independent of particular geoms and can be reused for generating plan fragments for any geom.

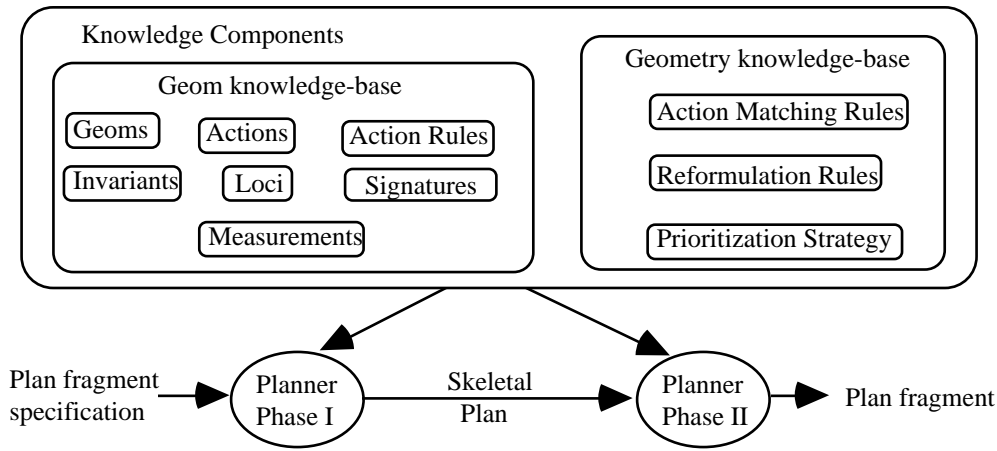

Figure 3. Architectural overview of the plan fragment generator

### 2.1 Geom Knowledge-base

The geom specific knowledge-base can be further decomposed into seven knowledge components.

### 2.1.1 ACTIONS

These describe operations that can be performed on geoms. In the GCE domain, three actions suffice to change the configuration of a body to an arbitrary configuration: *(translate g v)* which denotes a translation of geom *g* by vector *v*; *(rotate g pt ax amt)* which denotes a rotation of geom *g*, around point *pt*, about an axis *ax*, by an angle *amt*; and *(scale g pt amt)* where *g* is a geom, *pt* is a point on the geom, and *amt* is a scalar. The semantics of a scale operation depends on the type of the geom; for example, for a circle, a scale indicates a change in the radius of the circle and for a line-segment it denotes a change in the line-segment's length. *Pt* is the point on the geom that is fixed (e.g., the center of a circle).

### 2.1.2 INVARIANTS

These describe constraints to be solved for the geoms. The initial version of our system has been designed to generate plan fragments for a variable-radius *circle* and a variable length *line-segment* on a fixed workplane, with constraints on the distances between these geoms and points, lines, and other geoms on the same workplane. There are seven invariant types to represent these constraints. Examples of two such invariants are:

- *(Invariant-point g pt glb-coords)* which specifies that the point *pt* of geom *g* is coincident with the global coordinates *glb-coords,* and
- *(Fixed-distance-point g pt dist bias)* which specifies that the geom *g* lies at a fixed distance *dist* from point *pt*; *bias* can be either BIAS_INSIDE or BIAS_OUTSIDE depending on whether *g* lies inside or outside a circle of radius *dist* around point *pt*.

### 2.1.3 LOCI

These represent sets of possible values for a geom parameter, such as the position of a point on a geom. The various kinds of loci can be grouped into either a 1d-locus (representable by a set of parametric equations in one parameter) or a 2d-locus (representable by a set of parametric equations in two variables). For, example a line is a 1d locus specified as *(make-line-locus through-pt direc)* and represents an infinite line passing through *through-pt* and having a direction *direc*. Other loci represented in the system include rays, circles, parabolas, hyperbolas, and ellipses.

### 2.1.4 MEASUREMENTS

These are used to represent the computation of some function, object, or relationship between objects. These terms are mapped into a set of service routines which get called by the plan fragments. An example of a measurement term is: *(0d-intersection 1d-locus1 1d-locus2).* This represents the intersection of two 1d-loci. In the normal case, the intersection of two 1-dimensional loci is a point. However, there may be singular cases, for example, when the two loci happen to coincide; in such a case their intersection returns one of the locus instead of a point. There may also be degenerate cases, for example, when the two loci do not intersect; in such a case, the intersection is undefined. These exceptional conditions are also represented with each measurement type and are used during the second phase of the plan generation process to elaborate a skeletal plan (see Section 3.3).

### 2.1.5 GEOMS

These are the objects of interest in solving geometric constraint satisfaction problems. Examples of geoms are lines, line-segments, circles, and rigid bodies. Geoms have degrees of freedoms which allow them to vary in location and size. For example, in 3D-space a circle with a variable radius, has three translational, two rotational, and one dimensional degree of freedom.

The *configuration variables* of a geom are defined as the minimal number of real-valued parameters required to specify the geometric entity in space unambiguously. Thus, a circle has six configuration variables (three for the center, one for the radius, and two for the plane normal). In addition, the representation of each geom includes the following:

- *name*: a unique symbol to identify the geom;
- *action-rules*: a set of rules that describe how invariants on the geom can be preserved or achieved by actions (see below);
- *invariants*: the set of current invariants on the geom;
- *invariants-to-be-achieved*: the set of invariants that need to be achieved for the geom.

### 2.1.6 ACTION RULES

An action rule describes the effect of an action on an invariant. There are two facts of interest to a planner when constructing a plan: (1) how to achieve an invariant using an action and (2) how to choose actions that preserve as many of the existing invariants as possible. In general, there are several ways to achieve an invariant and several actions that will preserve an invariant. The intersection of these two sets of actions is the set of feasible solutions. In our system, the effect of actions is represented as part of geom-specific knowledge in the form of Action rules, whereas knowledge about how to compute intersections of two or more sets of actions is represented as geometry-specific knowledge (since it does not depend on the particular geom being acted on).

An action rule consists of a three-tuple *(pattern, to-preserve, to-[re]achieve)*. *Pattern* is the invariant term of interest; *to-preserve* is a list of actions that can be taken without violating the pattern invariant; and *to-[re]achieve* is a list of actions that can be taken to achieve the invariant or re-achieve an existing invariant "clobbered" by an earlier action. These actions are stated in the most general form possible. The matching rules in the Geometry Knowledge base are then used to obtain the most general unifier of two or more actions. An example of an action rule, associated with variable-radius circle geoms is:

> *pattern:* (1d-constrained-point ?circle (>> ?circle CENTER) ?1dlocus)          (AR-1)
> *to-preserve:* (scale ?circle (>> ?circle CENTER) ?any)
>         (translate ?circle (v- (>> ?1dlocus ARBITRARY-POINT)
>               (>> ?circle CENTER))
> *to-[re]achieve:* (translate ?circle (v- (>> ?1dlocus ARBITRARY-POINT)
>              (>> ?circle CENTER))

This action rule is used to preserve or achieve the constraint that the center of a circle geom lie on a 1d locus. There are two actions that may be performed without violating this constraint: (1) scale the circle about its center. This would change the radius of the circle but the position of the center remains the same and hence the 1d-constrained-point invariant is preserved. (2)

translate the circle by a vector that goes from its current center to an arbitrary point on the 1-dimensional locus (*(v- a b)* denotes a vector from point *b* to point *a*). To *achieve* this invariant only one action may be performed: translate the circle so that its center moves from its current position to an arbitrary position on the 1-dimensional locus.

### 2.1.7 SIGNATURES

For completeness, it is necessary that there exist a plan fragment for each possible combination of constraints on a geom. However, in many cases, two or more constraints describe the same situation for a geom (in terms of its degrees of freedom). For example, the constraints that ground the two end-points of a line-segment and the constraints that ground the direction, length, and one end-point of a line-segment both reduce the degrees of freedom of the line-segment to zero and hence describe the same situation. In order to minimize the number of plan fragments that need to be written, it is desirable to group sets of constraints that describe the same situation into equivalence classes and represent each equivalence class using a canonical form.

The state of a geom, in terms of the prior constraints on it, is summarized as a *signature*. A *signature scheme* for a geom is the set of canonical signatures for which plan fragments need to be written. In Kramer's earlier work (1993) the signature scheme had to be determined manually by examining each signature obtained by combining constraint types and designating one from a set of equivalent signatures to be canonical. Our approach allows us to construct the signature scheme for a geom automatically by using reformulation rules (described shortly). A reformulation rule rewrites one or more constraints into a simpler form. The signature scheme is obtained by first generating all possible combinations of constraint types to yield the set of all possible signatures. These signatures are then reduced using the reformulation rules until each signature is reduced to the simplest form. The set of (unique) signatures that are left constitute the signature scheme for the geom.

As an example, consider the set of constraint types on a variable radius circle. The signature for this geom is represented as a tuple *<Center, Normal, Radius, FixedPts, FixedLines>* where:

- *Center* denotes the invariants on the center point and can be either Free (i.e., no constraint on the center point), L2 (i.e., center point is constrained to be on a 2-dimensional locus), L1 (i.e., center point is constrained to be on a 1-dimensional locus), or Fixed.
- *Normal* denotes the invariant on the normal to the plane of the circle and can be either Free, L1, or Fixed (in 2D it is always fixed).
- *Radius* denotes the invariant on the radius and can be either Free or Fixed.
- *FixedPts* denotes the number of Fixed-distance-point invariants and can be either 0,1, or 2.
- *FixedLines* denotes the number of Fixed-distance-line invariants and can be either 0,1, or 2.

L2 and L1 denote a 2D and 1D locus respectively. If we assume a 2D geometry, the L2 invariant on the Center is redundant, and the Normal is always Fixed. There are then 3 x 1 x 2 x 3 x 3 = 54 possible signatures for the geom. However, several of these describe the same situation. For example, the signature:

*<Center-Free,Radius-Free, FixedPts-0,FixedLines-2>*

which describes a circle constrained to be at specific distances from two fixed lines, can be rewritten to:

*<Center-L1, Radius-Free,FixedPts-0,FixedLines-0>*

which describes a circle constrained to be on a 1-dimensional locus (in this case the angular bisector of two lines). Using reformulation rules, we can derive the signature scheme for variable radius circles consisting of only 10 canonical signatures given below:

<Center-Free,Radius-Free, FixedPts-0,FixedLines-0>
<Center-Free,Radius-Free, FixedPts-0,FixedLines-1>
<Center-Free,Radius-Free, FixedPts-1,FixedLines-0>
<Center-Free,Radius-Fixed, FixedPts-0,FixedLines-0>
<Center-L1,Radius-Free, FixedPts-0,FixedLines-0>
<Center-L1,Radius-Free, FixedPts-0,FixedLines-1>
<Center-L1,Radius-Free, FixedPts-1,FixedLines-0>
<Center-L1,Radius-Fixed, FixedPts-0,FixedLines-0>
<Center-Fixed,Radius-Free, FixedPts-0,FixedLines-0>
<Center-Fixed,Radius-Fixed, FixedPts-0,FixedLines-0>

Similarly, the number of signatures for line-segments can be reduced from 108 to 19 using reformulation rules.

## 2.2 Geometry Specific Knowledge

The geometry specific knowledge is organized as three different kinds of rules.

### 2.2.1 MATCHING RULES

These are used to match terms using geometric properties. The planner employs a unification algorithm to match actions and determine whether two actions have a common unifier. However, the standard unification algorithm is not sufficient for our purposes, since it is purely syntactic and does not use knowledge about geometry. To illustrate this, consider the following two actions:

*(rotate $g $pt1 ?vec1 ?amt1),* and
*(rotate $g $pt2 ?vec2 ?amt2).*

The first term denotes a rotation of a fixed geom $g, around a fixed point $pt1 about an arbitrary axis  by an arbitrary amount. The second term denotes a rotation of the same geom around a different fixed point $pt2 with the rotation axis and amount being unspecified as before. Standard unification fails when applied to the above terms because no binding of variables makes the two terms syntactically equal[2]. However, resorting to knowledge about geometry, we can match the two terms to yield the following term:

*(rotate $g $pt1 (v- $pt2 $pt1) ?amt1)*

which denotes a rotation of the geom around the axis passing through points $pt1 and $pt2. The point around which the body is rotated can be any point on the axis (here it is arbitrarily chosen to be one of the fixed points, $pt1) and the amount of rotation can be anything.

The planner applies the matching rules to match the outermost expression in a term first; if no rule applies, it tries subterms of that term, and so on. If none of the matching rules apply, then this algorithm degenerates to standard unification. The matching rules can also have conditions attached to them. The condition can be any boolean function; however, for the most part they tend to be simple type checks.

---

[2] Specifically, unification fails when it tries to unify $pt1 and $pt2.

## 2.2.2 REFORMULATION RULES

As mentioned earlier, there are several ways to specify constraints that restrict the same degrees of freedom of a geom. In GCE, plan fragments are indexed by signatures which summarize the available degrees of freedom of a geom. To reduce the number of plan fragments that need to be written and indexed, it is desirable to reduce the number of allowable signatures. This is accomplished with a set of invariant reformulation rules which are used to rewrite pairs of invariants on a geom into an equivalent pair of simpler invariants (using a well-founded ordering). Here equivalence means that the two sets of invariants produce the same range of motions in the geom. This reduces the number of different combinations of invariants for which plan fragments need to be written. An example of invariant reformulation is the following:

(fixed-distance-line ?c ?l1 ?d1 BIAS_COUNTERCLOCKWISE)
(fixed-distance-line ?c ?l2 ?d2 BIAS_CLOCKWISE)

$$\Downarrow \qquad\qquad\qquad\qquad\qquad \text{(RR-1)}$$

(1d-constrained-point ?c (>> ?c center) (angular-bisector
                                    (make-displaced-line ?l1 BIAS_LEFT ?d1)
                                    (make-displaced-line ?l2 BIAS_RIGHT ?d2)
                                    BIAS_COUNTERCLOCKWISE
                                    BIAS_CLOCKWISE))

This rule takes two invariants: (1) a geom is at a fixed distance to the left of a given line, and (2) a geom is at a fixed distance to the right of a given line. The reformulation produces the invariant that the geom lies on the angular bisector of two lines that are parallel to the two given lines and at the specified distance from them. Either of the two original invariants in conjunction with the new one is equivalent to the original set of invariants.

Besides reducing the number of plan fragments, reformulation rules also help to simplify action rules. Currently all action rules (for variable radius circles and line-segments) use only a single action to preserve or achieve an invariant. If we do not restrict the allowable signatures on a geom, it is possible to create examples where we need a sequence of (more than one) actions in the rule to achieve the invariant, or we need complex conditions that need to be checked to determine rule applicability. Allowing sequences and conditionals on the rules increases the complexity of both the rules and the pattern matcher. This makes it difficult to verify the correctness of rules and reduces the efficiency of the pattern matcher.

Using invariant reformulation rules allows us to limit action rules to those that contain a single action. Unfortunately, it seems that we still need conditions to achieve certain invariants. For example, consider the following invariant on a variable radius circle:

      (fixed-distance-point ?circle ?pt ?dist BIAS_OUTSIDE)

which states that a circle, *?circle* be at some distance *?dist* from a point *?pt* and lie outside a circle around *?pt* with radius *?dist*. One action that may be taken to achieve this constraint is:

     (scale  ?circle
              (>> ?circle center)

```
(minus (>> (v- (>> ?circle center) ?pt)
                magnitude)
        ?dist)
```

that is, scale the circle by setting its radius to the distance between its center and the point *?pt* minus the scalar amount *?dist* (see Figure 4). However, this action achieves the constraint only when the circle happens to lie outside the circular region of radius *?dist* and center *?pt*.

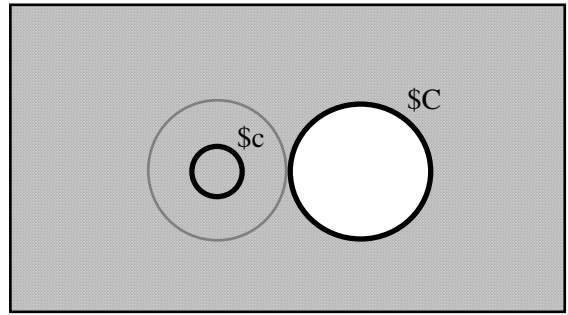

Figure 4. The geom $c can be scaled to touch $C only if the center of $c lies in the shaded region.

Therefore, we need a pre-condition to the rule that checks if this is indeed the case. Note that the above action is necessary for completeness (otherwise the planner would not be able to solve certain cases which have a solution). Instead of allowing conditional rules, we use rules without condition and in the second phase of the plan generation check to see that there are no exceptions. Thus, in the above example, an exception would be detected since the third argument of the scale operation returns a negative number – considered an exception condition for a scale operation.

### 2.2.3 PRIORITIZING STRATEGY

Given a set of invariants to be achieved on a geom, a planner generally creates multiple solutions. All of these are valid solutions and in the absence of exception conditions will yield the same configuration of a geom. However, some plan fragments will contain redundant action sequences (e.g., two consecutive translations). Moreover, when the geom is under constrained or when there are exception conditions, some plan fragments will be able to provide a solution whereas others will not. The prioritization strategy is used to prioritize the skeletal plan fragments so that plan fragments with the least redundancy and most flexibility can be chosen.

Eliminating plan fragments with redundant actions turns out to be straightforward. We assume that there is only one degree of dimensional freedom for each geometric body. Under this assumption it can be proved that 1 translation, 1 rotation, and 1 scale is sufficient to change the configuration of an object to an arbitrary configuration in 3D space. Therefore, any plan fragment that contains more than one instance of an action type contains redundancies and can be rewritten to an equivalent plan fragment by eliminating redundant actions, or combining two or more action into a single composite action. As an example, consider the following pair of translations on a geom:

• (translate $g ?vec)

- (translate $g (v- ?to$_2$ (>> $g center)))

where *?vec* represents an arbitrary vector and *?to*$_2$ represents an arbitrary position. If *?to*$_2$ is independent of any positional parameter of the geom, then the first translate action is redundant and can be removed. Hence all plan fragments that contain such redundant actions can be eliminated.

To prioritize the remaining plan fragments the following principle is used:

> *Prefer solutions that subsume an alternative solution.*

The rationale for this principle is that it permits greater flexibility in solving constraints when there are exception conditions. For example, suppose there are two solutions for a circle geom:

> *Solution 1:* Translate the circle so that the center lies at a fixed position on a 1-dimensional locus.

> *Solution 2:* Translate the circle so that the center lies at an arbitrary point on a 1-dimensional locus; then scale by some fixed amount (which is a function of the position of the arbitrary point).

The first solution is subsumed by the second solution since we can always choose the arbitrary point in *Solution 2* to be at the fixed position specified in *Solution 1* (the scale operation in that case leaves the dimension of the circle unchanged). Therefore Solution 2 is preferred over Solution 1.

The subsumption relation imposes a partial order on the set of skeletal plan fragments. The prioritization strategy selects the maximal elements of this partial order. At runtime each of these is tried in turn until one of them yields a solution.

## 3.0 Plan Fragment Generation

The plan fragment generation process is divided into two phases (Figure 1). In the first phase a specification of the plan fragment is taken as input, and a planner is used to generate a set of skeletal plans. These form the input to the second phase which chooses one or more of the skeletal plans and elaborates them to take care of singularities and degeneracies. The output of this phase are complete plan fragments.

### 3.1 Phase I

A skeletal plan is generated using a breadth-first search process. Figure 5 gives the general form of a search tree produced by the planner. The first action is typically a reformulation where the planner uses the reformulation rules to rewrite the geom invariants into a canonical form. Next, the planner searches for actions that produce a state in which at least 1 invariant in the *Preserved* list is preserved or at least 1 action in the *To-be-achieved* (*TBA*) list is achieved. The preserved and achieved invariants are pushed into the *Preserved* list, and the clobbered or unachieved invariants are pushed into the *TBA* list of the child state.

The above strategy will produce intermediate nodes in the search tree which might clobber one or more preserved invariant without achieving any new invariant or might produce a state which is identical to its parent state in terms of the invariants on the Preserved and TBA list. This is because in the initial state a geom may be in some arbitrary configuration (among a set of allowable configurations) and it may be necessary to first move the geom to an alternative allowable configuration to find the optimal solution.

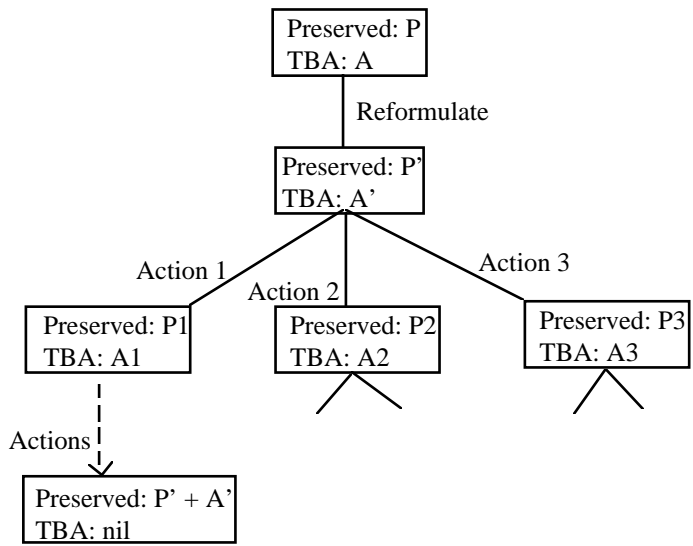

Figure 5. Overview of the search tree produced by the planner

To illustrate this need, consider the example in Figure 6. In this example, there is one prior constraint on the variable radius circle geom: its center lies on a 1-dimensional locus. The new constraint to be achieved is: the geom should lie at a fixed distance from a line. In order to *achieve* this constraint only one of the following two actions may be taken: 1) scale the circle so that it is at a fixed distance from the line (Figure 6b), or 2) translate the circle to a new position on the 1-dimensional locus so that it touches the line (Figure 6c). However, there are an infinite number of additional solutions consisting of combinations of scale and translation (Figure 6d). These solutions can be derived if the planner first changes the configuration of the geom so that it only preserves the existing invariant without achieving the new invariant (i.e., scale by an arbitrary amount or translate to an arbitrary point on the 1-dimensional locus) followed by an action that achieves the new invariant. Therefore the planner also creates child states that are identical to the parent state in terms of invariants on the *Preserved* and *TBA* lists.

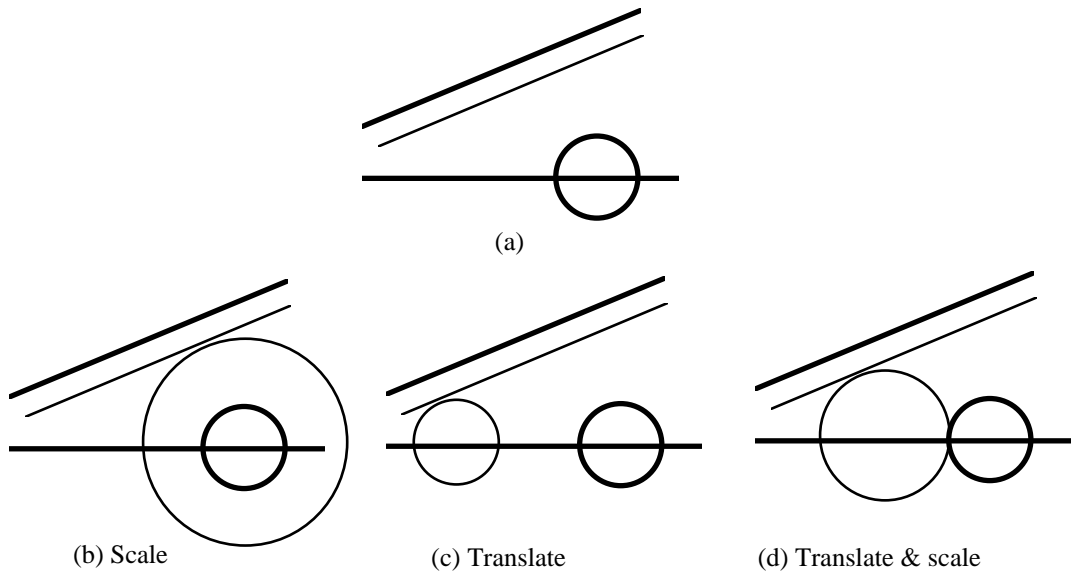

Figure 6. Example to illustrate the need for actions that produce a state equivalent to the parent state.

The planner iteratively expands each leaf node in the search tree until one of the following is true:

1. The node represents a solution; that is, the *TBA* list is nil.
2. The node represents a cycle; that is, the invariants in the *Preserved* and *TBA* lists are identical to one of the ancestor nodes.

The node is then marked as terminal and the search tree is pruned at that point. If all leaf nodes are marked as terminal, then the search terminates. The planner then collects all terminal nodes that are solutions. The plan-steps of each of those solution nodes represents a skeletal plan fragment. When multiple skeletal plan fragments are obtained by the planner, one of them is chosen using the prioritizing rule described earlier and is passed to the second phase of the plan fragment generation.

### 3.2 Phase I: Example

We use the example of Section 1 to illustrate Phase I of the planner. The planner begins by attempting to reformulate the given constraints. It uses reformulation rule RR-1 described earlier and repeated below for convenience:

(fixed-distance-line ?c ?l1 ?d1 BIAS_COUNTERCLOCKWISE)
(fixed-distance-line ?c ?l2 ?d2 BIAS_CLOCKWISE)

$$\Downarrow$$ (RR-1)

(1d-constrained-point ?c (>> ?c center) (angular-bisector

(make-displaced-line ?l1 BIAS_LEFT ?d1)
(make-displaced-line ?l2 BIAS_RIGHT ?d2)
BIAS_COUNTERCLOCKWISE
BIAS_CLOCKWISE))

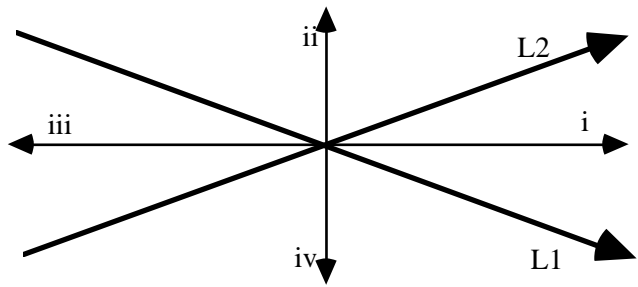

Figure 7. Four possible angular bisectors of two lines L1 and L2. The bias symbols for L1 and L2 corresponding to ray (i) is BIAS_COUNTERCLOCKWISE & BIAS_CLOCKWISE respectively.

In the above rule there are two measurement terms: *make-displaced-line* and **angular-bisector**. *Make-displaced-line* takes three arguments: a line, *l*, a bias symbol indicating whether the displaced line should be to the left or right of *l*, and a distance, *d*. It returns a line parallel to the given line *l* at a distance *d* to the left or right of the line depending on the bias. *Angular-bisector* takes two lines, *l1* and *l2*, and two bias symbols and returns one of the four rays that bisects the lines *l1* and *l2* depending on the bias symbols (see Figure 7). After reformulation, the state of the search tree is as shown in Figure 8. No further reformulation rules are applicable at this point.

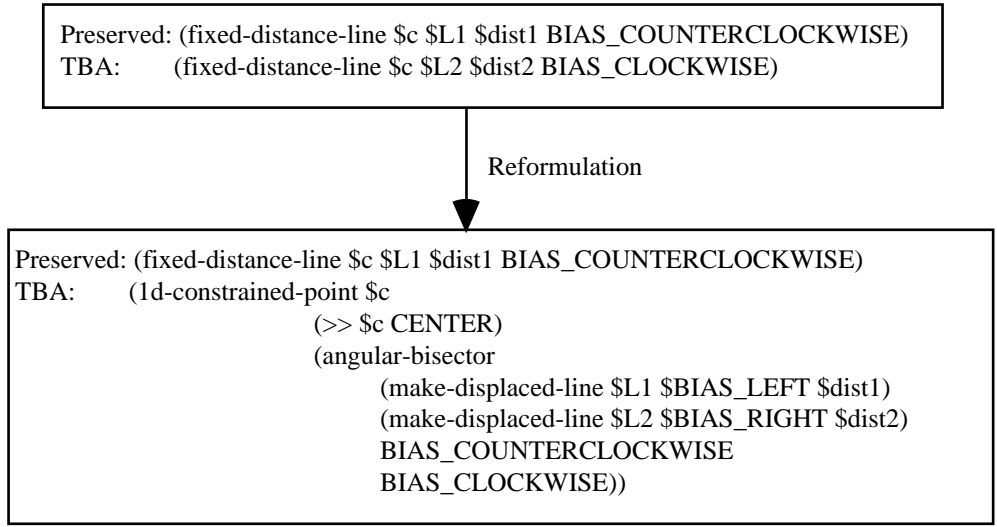

Figure 8. Search tree after reformulating invariants

Next, the planner searches for actions that can achieve the new invariant or preserve the existing invariant or do both. We only describe the steps involved in finding actions that satisfy the maximal number of constraints (in this case, two). The planner first finds all actions that achieve the *1d-constrained-point* invariant by examining the action rules associated with the variable-circle geom. The action rule AR-1 contains a pattern that matches the *1d-constrained-point* invariant:

    *pattern:* (1d-constrained-point ?circle (>> ?circle center) ?1dlocus)       (AR-1)
    *to-preserve:* (scale ?circle (>> ?circle center) ?any)
             (translate ?circle (v- (>> ?1dlocus arbitrary-point)
                          (>> ?circle center))
    *to-[re]achieve:* **(translate ?circle (v- (>> ?1dlocus arbitrary-point)**
                            **(>> ?circle center))**

with the following bindings:

    {?circle = $c, ?1d-locus = (angular-bisector (make-displaced-line ...) ...)}

Substituting these bindings we obtain the following action:

```
(translate $c (v- (>> (angular-bisector (make-displaced-line $L1 BIAS_LEFT $dist1)
                                        (make-displaced-line $L2 BIAS_RIGHT $dist2)
                      arbitrary-point)
                  (>> $c center)))                                                    (a1)
```

which can be taken to achieve the constraint. Similarly, the planner finds all actions that will preserve the fixed-distance-line invariant. The relevant action rule is the following:

```
pattern:      (fixed-distance-line ?circle ?line ?distance)                          (AR-2)
to-preserve:  (translate ?circle (v-   (>> (make-line-locus (>> ?circle center)
                                                            (>> ?line direction))
                           arbitrary-point)
                       (>> ?circle center))
to-[re]achieve: (translate ?circle (v- (>> (make-displaced-line
                                              ?line
                                              BIAS_LEFT
                                              (plus ?distance (>> ?circle radius)))
                           arbitrary-point)
                       (>> ?circle center)))
```

The relevant action after the appropriate substitutions is:

```
(translate $c (v- (>> (make-line-locus
                          (>> $c center)
                          (>> L1 direction))
                  arbitrary-point)
              (>> $c center))                                                         (a2)
```

Now, to find an action that both preserves the preserved invariant and achieves the TBA invariant, the planner attempts to match the preserving action (a2) with the achieving action (a1). The two actions do not match using standard unification, but match employing the following geometry-specific matching rule:

```
(v- (>> $1d-locus1 arbitrary-point) $to)          # To move to an arbitrary point on two
(v- (>> $1d-locus2 arbitrary-point) $to)          # different loci, move to the point that
            ⇓                                      # is the intersection of the two loci
(v- (0d-intersection $1d-locus1 $1d-locus2) $to)
```

to yield the following action:

```
(translate $c (v- (0d-intersection (angular-bisector
                                       (make-displaced-line ...) ...)
                    (make-line-locus (>> $c center) (>> $L1 direction)))
(>> $c CENTER)))
```

This action moves the circle to the point shown in Figure 9 and achieves both the constraints. This simple one-step plan constitutes a skeletal plan fragment.

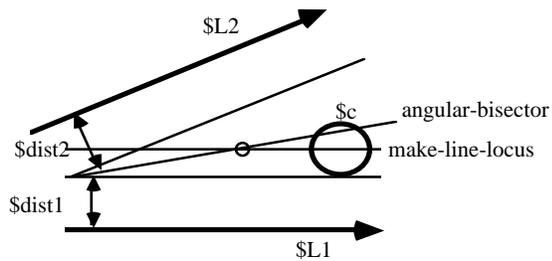

Figure 9. The ○ denotes the point to which the circle is moved.

There are two other actions that are generated by the planner in the first iteration. One of these achieves the new constraint but clobbers the prior invariant. The other moves the circle to another configuration without achieving the new constraint but preserving the prior constraint. The first action produces a terminal state since there are no more constraints to be achieved. Hence the search tree is pruned at that point. However, the planner continues to search for alternative solutions by expanding the other two nodes. After two iterations the following solutions are obtained:

1. Translate to the intersection of the *angular-bisector* and *make-line-locus*.
2. Translate to an arbitrary point on the *angular-bisector*, followed by a translation to the intersection point.
3. Translate to an arbitrary point of *make-line-locus*, followed by a translation to the intersection point.
4. Translate to an arbitrary point on the *angular-bisector* and then scale.

At this stage the first phase of the plan fragment generation is terminated and the skeletal plan fragments are passed on to the second phase of the planner.

### 3.3 Phase II: Elaboration of Skeletal Plan Fragment

The purpose of Phase 2 planning is to i) select one or more skeletal plan fragments, and ii) elaborate them so that they generate the most desirable configuration when the geom is under constrained as well as handle exception conditions.

#### 3.3.1 SELECTION OF SKELETAL PLAN FRAGMENTS

There are two primary considerations in selecting a skeletal plan fragment – reduce redundant actions in the plan and increase generality of the plan. These considerations are used to formulate a prioritization strategy described in Section 2. The strategy is implemented as a lookup table that assigns weights to the various plan fragments. The plan fragments with the maximal weights are selected for elaboration by Phase 2. Readers interested in the implementation details are referred to (Hoar, 1995).

#### 3.3.2 PLAN FRAGMENT ELABORATION

Plan fragment elaboration refines a skeletal plan fragment in two ways. First, it refines actions that are under constrained (e.g., translate to an arbitrary point on a locus) by appropriate instantiation of the unconstrained parameters (e.g., selecting a specific point on a locus). Second, it handles exception conditions that result in under constrained or over-constrained systems. Both action refinement and exception handling are treated using a common technique.

Plan elaboration is based on the "principle of least motion": when there are multiple solutions for a problem choose the solution that minimizes the total amount of perturbation (motion) in the system. Implementing the principle requires the definition of a motion function, $C_{A,G}$ for each action, A, and geom type, G. For example, for a translation of a geom, the motion function, $C_{T,circle}$ could be the square of the displacement of the center of the geom from its initial to its final position. We also need a motion summation function, $\Sigma_G$ that sums the motion produced by individual actions on a geom G. An example of the summation function is the normal addition operator: *plus*. The total motion produced in a geom is computed using the summation function and the motion functions for action-geom pairs.

When a plan fragment is under constrained, the expression representing the total motion would contain one or more variables representing the ungrounded parameters of the geom. Formal optimization techniques, based on finite difference methods, can be used to obtain values of the parameters that would minimize the motion function. However, we use a more efficient, algorithm based on hill-climbing which does not guarantee optimality but yields good results in practice. The use of this heuristic algorithm is justified in many interactive applications like sketching, where a fast, sub-optimal solution is preferable to a computationally expensive, optimal one.

 The algorithm begins by segmenting all continuous loci into discrete intervals. It then systematically searches the resultant, discrete n-dimensional space. The algorithm first finds a local minima along one dimension while holding the other variables at constant values. Then it holds the first variable at the minimum value found and searches for a lower local minima along the second dimension and so on. Although this algorithm does not guarantee finding a global or even a local minima, it is very efficient and yields good results in practice. The implemented algorithm is somewhat more complex than the simple description above; further details can be found elsewhere (Hoar, 1995).

Exception conditions can be handled using the same technique as above. Exception conditions are identified when a service routine returns a set of solutions or no solution (e.g., a routine to compute the intersection of two 1-dimensional loci returns a 1-dimensional locus or nil). Multiple solutions represent an under constrained system and requires a search among the set of solutions returned. These conditions are handled exactly as described in the previous paragraph. When a no-solution exception occurs, the system aborts the plan fragment and prints a diagnostic message explaining why the constraint could not be solved.

### 3.4 Phase II: Example

Four skeletal plan fragments were generated in the first phase of the planner (Section 3.2). Using the rule for eliminating redundant translations given earlier, the second and third plan fragments can be reduced to single translation plan fragments equivalent to the first plan fragment. This leaves only two distinct plan fragment solutions to consider.

Using the prioritizing rule, the system concludes that the first plan fragment consisting of a single translation is subsumed by the second plan fragment consisting of a translation and a scale. Thus, the second plan fragment is chosen as the preferred solution.

This plan fragment is not deterministic since it contains an action that translates the circle geom to an arbitrary point on the angular-bisector. Therefore, the system inserts an iterative loop that computes the amount of motion of the circle for various points on the angular bisector, breaking out of the loop when it finds a minima. Similarly, for each service routine that may return an exception, the system inserts a case statement which contains a loop to handle situations when more than one solution is returned. Online Appendix 1 contains a complete example of a plan fragment generated by the system.

## 4.0 Results

The plan fragment generator described here has been implemented using CLOS (Common Lisp Object System). We have implemented parts of the geometric constraint engine (GCE) described by Kramer in C++ with an XMotif based graphical user interface. We have also written a translator that translates the synthesized plan fragments into C++. A complete plan fragment library for a representative geom (line segment) has been synthesized and integrated with the constraint engine. Using this we have been able to successfully demonstrate the solution of several geometric constraints. We present below an evaluation of the system.

The primary contribution of this research is not a novel geometric constraint satisfaction approach. From the perspective of constraint satisfaction techniques, the novel feature of our approach - degrees of freedom analysis - has already been described in earlier works by the second author (Kramer, 1992, 1993). The goal of this research was to develop automated techniques that will enable the degrees of freedom approach to scale up by reducing the amount of effort needed in creating plan fragment libraries. Hence, our evaluation is based on how successful we have been in automating the plan fragment synthesis process.

We have used the plan fragment generator described above to automatically synthesize plan fragments for two representative geoms -- line-segments and circles -- in 2D. There are seven types of constraints and thirty four rules in the system (12 action rules for line-segments, 8 action rules for circles, 7 Reformulation rules, and 7 Matching rules). Using these rules we have successfully generated skeletal plan fragments for various combinations of constraints on line segments (249) and circles (50). The largest search tree produced by the planner is on the order of a few hundred nodes and takes a few minutes on a Macintosh Quadra. For evaluation purposes, we present data for one representative geom - line segment.

### 4.1 Programming Effort

Figure 10 shows the number of lines of code comprising the current system. The areas in solid represent code that was written manually. This includes about 5000 lines of CLOS code for the plan fragment synthesizer, 5400 lines of C/C++ for the user interface, and 3300 lines of C/C++ for the support routines. The hatched area represent code that was synthesized by the plan fragment generator. It represents about 27000 lines of C++ code (for plan fragments for the line-segment geom). The size of the synthesized plan fragment (about 121 lines average) is much less than that of plan fragments written manually (in C) in the original version of GCE. Thus, using an automated plan fragment generator has considerably reduced the amount of programming. While a reduction ratio of 5:1 is a good indicator of the reduction in programming effort, it is subject to criticism since it compares code in two very different programming languages and comprising different degrees of difficulty.

A more accurate evaluation is obtained by comparing the total effort required in writing plan fragments manually against the total effort required in synthesizing them using the technique described in this paper. It is extremely difficult, if not impossible, to do this in any controlled experimental setting because of the number of factors and cost involved. The best that can be done is to compare the empirical data based on our experience in developing the system. The following table shows the effort in person days in developing the plan fragment library for the line-segment geom using our technique.

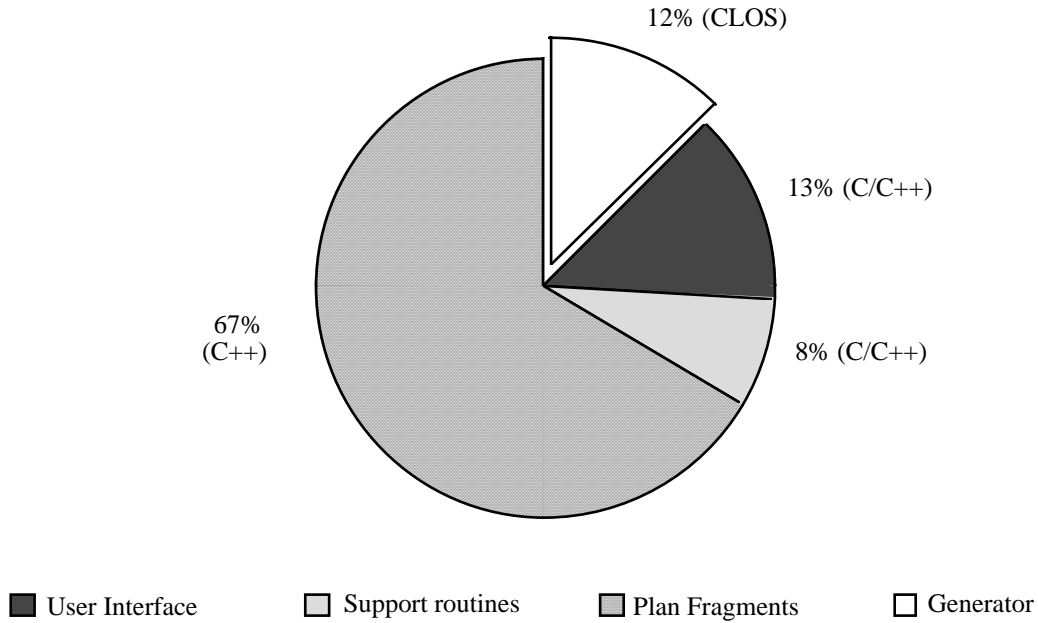

Figure 10. Lines of code in different parts of the system

|  | Research | Development | Total |
|---|---|---|---|
| Plan Fragment Generator | 90 | 150 | 210 |
| Manually | 0 | *498* | *498* |

**Table 1.** Effort (in person-days) in creating plan fragments

For the effort involved in writing plan fragments manually, we use a conservative estimate of 2 person days for each plan fragment[3]. The table shows that using the plan fragment generator we obtained a 58% reduction in effort in creating the plan fragment library. The testing and debugging time has been ignored and assumed to be the same for both cases (although we believe that this time is much more for manually generated plan fragments).

### 4.2 Scalability

A much stronger evidence in support of our technique is obtained when we look at the effort required in extending the plan fragment library by adding more features (e.g., new kinds of geoms or constraints). To evaluate the scalability of the approach, we decided to extend the plan fragments to 3D where geoms have added degrees of rotational and translational freedom. Such an extension when done manually would be significant exercise in software maintenance since it requires changes to each plan fragment in the library. Using the plan fragment generator we only needed to revise the rules used by the planner and make changes to the support routines. Since the support routines were written manually, the cost to modify them is the same in both approaches, and only the effort needed to rewrite the rules is relevant. *It took only 1 week of effort to rewrite and debug the action rules and synthesize the complete plan fragment library for 3D, and link it successfully with the constraint engine.* This is a significant result in demonstrating that our technique can be used to scale up degrees of freedom analysis to more complex geoms and geometries.

---

[3] This estimate is based both on the effort required in developing the plan fragment library for GCE as well as experimental data obtained by having two graduate students write a few plan fragments manually.

### 4.3 Correctness

An important issue that has been ignored so far is: how does one verify the correctness and completeness of the plan fragment generator? We have done extensive testing and evaluation of the plan fragments synthesized by the plan fragment generator. Table 2 summarizes the results.

| Number of plan fragment specifications | | 249 |
|---|---|---|
| **Completeness** | Specs. with no solutions: | |
| |    No solution exists | 65 |
| |    Missing rules | 13 |
| |    No symbolic solution | 2 |
| |    Total | 80 |
| **Correctness** | Plan fragments with errors: | |
| |    Faults due to errors in logic | 0 |
| |    Support routine errors | 56 |
| |    Total | 56 |

**Table 2.** Completeness and Correctness of synthesized plan fragments

There were eighty plan fragment specifications for which the planner failed to produce a solution. In sixty five of these specifications, there were no solutions in the general case -- these specifications represent overconstrained problems, such as constraining one end point of a line-segment to be on a one-dimensional locus when previous constraints have already reduced that end-point's translational degrees of freedom to zero. The only action the planner can take in such cases is to check that the new constraint is already satisfied. Thirteen of the cases had no solutions because of two missing rules: one action rule, and one reformulation rule. Once the two rules were added all the thirteen specifications were solved. Finally, there were only two plan fragments for which the planner failed to produce an analytical solution. The cases are shown in Figure 11. To solve such problems we need a reformulation rule that reformulates the existing invariant to a constraint that the endpoint of $lseg is the curve $L3. Instead of representing complex 1-dimensional (and higher dimensional) loci like $L3, we assume that the constraint engine would call a numerical solver that computes the solution iteratively. An alternative would be to extend the set of support routines to handle such complex loci and their intersections.

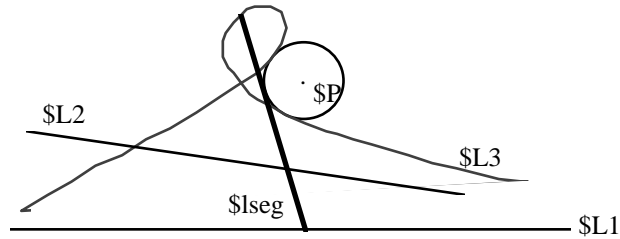

Figure 11. Example of problem that generated no symbolic solution. $lseg is a line-segment which is constrained to have one end-point on $L1, have a fixed length, and be tangent to a circle centered at $P. The new constraint is that the other end-point of $lseg be on $L2.

To check for the correctness of plan fragments, we did an exhaustive evaluation of all the plan fragments. As can be seen from Table 2, the code that has been synthesized is not perfect. About 20% of the plan fragments do not function correctly. We analyzed the reasons for the failure by manually inspecting the plan fragments. The most significant finding was that none of the failures were due to logical errors in the plan fragments. In other words the skeletal plan fragments being generated by Phase I were correct and complete. Most of the failures were because of bugs in the mathematical support routines called by the plan fragments. In a few instances the failures were traced to bugs in implementing Phase 2 of the plan fragment: either selecting the wrong skeletal plan fragment or not computing the least motion correctly. We had not expected the first version of the automatically generated plan fragments to be completely bug-free. Indeed, the high percentage of plan fragments that do function correctly (almost 80%) is a very positive result and reflects a significant increase in quality and a corresponding decrease in maintenance effort for building geometric constraint satisfaction systems using our approach.

## 5.0 Related Work

Geometric constraint satisfaction is an old problem. Probably the first application of this problem to constraint-based sketching was the Sketchpad program developed by Sutherland (1963). The Sketchpad program was based on constraint relaxation and was limited to problems that were modeled with point variables.

In the field of mechanical design, a graph based approach to constraint satisfaction has been described by Serrano (1987). In Serrano's approach the constraints are modeled using a constraint network; a constraint satisfaction engine finds the values of constrained variables that satisfy the constraints in the network using constraint propagation techniques. The approach identifies loops or cycles in the network, collapses them into supernodes, and then applies conventional sequential local propagation. This approach uses numerical iterative techniques which can have problems with stability. The computational advantage of this approach reduces when equations are tightly coupled.

Most of the commercial systems that do kinematics analysis are based on numerical iterative techniques or algebraic techniques or a combination of the two. Although these approaches are in principle robust, they have several shortcomings that make them inappropriate for real-time applications.

Among non-commercial systems, a notable new approach to constraint based sketching is the Juno-2 being developed at DEC-SRC (Heydon & Nelson, 1994). Constraints in Juno-2 are specified using an expressive, declarative constraint language which seems powerful enough to express most constraints that arise in practice. Juno-2 uses a combination of symbolic and numerical techniques to solve geometric constraints efficiently. A key difference between Juno-2 and the degrees of freedom approach is that in Juno-2 the symbolic reasoning is done in the domain of equations. For example, Juno-2 uses symbolic techniques like local propagation, unpacking, and unification closure to reduce the number of unknowns in a system of equations. The equations are then solved by Newton's method. In degrees of freedom analysis, the symbolic reasoning is done in the domain of geometry rather than equations.

Geometric constraints also arise in robotics, where the primary issues are concerned with finding a physically realizable path through space for a robot manipulator or a part of an assembly. A fundamental analytical tool for solving motion planning problems in robotics is the configuration space framework (Lozano-Perez, 1983). In configuration space approach, the problem of planning the motion of a part through a space of obstacles is transformed into an equivalent but simpler problem of planning the motion of a point through a space of enlarged configuration-space obstacles. Degrees of freedom analysis finesses this problem since it uses the notion of incremental assembly *only as a metaphor* for solving geometric constraint systems. No physical meaning is ascribed to how objects move from where they are to where they need to be - a factor that is quite important in a real-world assembly problem arising in robotics. The only use of the plan is to guide the solution of the complicated non-linear equations arising from formulating and solving the problems algebraically.

## 6.0 Conclusions

We have described a plan fragment generation methodology that can synthesize plan fragments for a geometric constraint satisfaction systems by reasoning from first principles about geometric entities, actions, and topology. The technique has been used to successfully synthesize plan fragments for a realistic set of constraints and geoms. It may seem that we have substituted one hard task - writing a complete set of correct plan fragments for various combinations of geoms and constraints - by an even harder task: creating the knowledge base of rules to automate the process. The rules *are* difficult to write and we have found that it is necessary to spend some effort in debugging the rules. However, we estimate that the total effort to write and debug rules is still an order of magnitude less than writing and debugging manually written plan fragment code. Our future work is to investigate how this approach scales up to more complex constraints and geometries.

Another useful extension of this work would be concerned with pushing the automation one level further so as to automatically acquire some types of knowledge from simpler building blocks. For example, a technique for automatically synthesizing the least motion function from some description of the geometry would be very useful.

In our method the plan fragment generation is divided into two disjoint phases. An alternative method would be to explore how the two phases can be interleaved. One possibility is that when there is a degeneracy because of a redundant constraint, the planner could reformulate the problem by removing the redundant constraint and re-synthesize a skeletal plan fragment with the new set of constraints. The resultant plan would form a part of the original plan fragment to deal with the degenerate cases. In other words, plan fragments would be generated on-the-fly as needed by the constraint solver.

## Acknowledgments

We thank Qiqing Xia who helped in implementing parts of the system described in this paper. We also acknowledge the support and resources provided by the School of Electrical Engineering and Computer Science, Washington State University. This work originated while the first author was at the Knowledge Systems Laboratory, Stanford University, and the second author was at the Schlumberger Laboratory of Computer Science, Austin.